\documentclass[11pt,a4paper]{article}
\usepackage{CJKutf8}
\usepackage[hyperref]{emnlp2021}
\usepackage{times}
\usepackage{latexsym}
\usepackage[T1]{fontenc}
\usepackage[utf8]{inputenc}
\usepackage{microtype}
\usepackage{tikz}
\usepackage{booktabs}
\usepackage{bm}
\usepackage{amssymb}
\usepackage{amsmath}
\usepackage{adjustbox}
\usepackage{listingsutf8}
\usepackage{multirow}
\usepackage{supertabular}
\usepackage{url}

\definecolor{codegreen}{rgb}{0,0.6,0}
\definecolor{codegray}{rgb}{0.5,0.5,0.5}
\definecolor{codepurple}{rgb}{0.58,0,0.82}
\definecolor{backcolour}{gray}{0.95}

\lstdefinestyle{ptm_code_style}{
  backgroundcolor=\color{backcolour},
  commentstyle=\color{codegreen},
  keywordstyle=\color{magenta},
  numberstyle=\tiny\color{codegray},
  stringstyle=\color{codepurple},
  basicstyle=\ttfamily\small,
  breakatwhitespace=false,         
  breaklines=true,                 
  captionpos=b,                    
  keepspaces=true,                 
  numbers=none,                    
  numbersep=5pt,                  
  showspaces=false,                
  showstringspaces=false,
  showtabs=false,                  
  tabsize=2,
  frame=tlbr,
  framesep=4pt,
  framerule=0pt,
  upquote=true
}

\title{N-LTP: An Open-source Neural Language Technology Platform \\ for Chinese}
\author{Wanxiang Che, Yunlong Feng, Libo Qin, Ting Liu \\
	Research Center for Social Computing and Information Retrieval \\
	Harbin Institute of Technology, China \\
	{\tt \{car,ylfeng,lbqin,tliu\}@ir.hit.edu.cn}	
}

\date{}

\begin{document}

\pgfdeclarelayer{background}
\pgfdeclarelayer{foreground}
\pgfsetlayers{background,main,foreground}


\tikzstyle{sensor}=[draw, fill=blue!20, text width=5em,
text centered, simum height=2.5em,drop shadow]
\tikzstyle{ann} = [above, text width=5em, text centered]
\tikzstyle{wa} = [sensor, text width=10em, fill=red!20,
minimum height=6em, rounded corners, drop shadow]
\tikzstyle{sc} = [sensor, text width=13em, fill=red!20,
minimum height=10em, rounded corners, drop shadow]

\def\blockdist{2.3}
\def\edgedist{2.5}

\maketitle
\begin{abstract}
  We introduce \texttt{N-LTP}, an open-source neural language technology platform supporting six fundamental Chinese NLP tasks: {lexical analysis} (Chinese word segmentation, part-of-speech tagging, and named entity recognition), {syntactic parsing} (dependency parsing), and {semantic parsing} (semantic dependency parsing and semantic role labeling).
  Unlike the existing state-of-the-art toolkits, such as \texttt{Stanza}, that adopt an independent model for each task, \texttt{N-LTP} adopts the multi-task framework by using a shared pre-trained model, which has the advantage of capturing the shared knowledge across relevant Chinese tasks. In addition,
  a knowledge distillation method \cite{DBLP:journals/corr/abs-1907-04829} where the single-task model teaches the multi-task model is further introduced to encourage the multi-task model to surpass its single-task teacher.
  Finally, we provide a collection of easy-to-use APIs and a visualization tool to make users to use and view the processing results more easily and directly.
  To the best of our knowledge, this is the first toolkit to support six Chinese NLP fundamental tasks.
  Source code, documentation, and pre-trained models are available at \url{https://github.com/HIT-SCIR/ltp}.
\end{abstract}

\begin{table*}[htbp]
  \centering
  \begin{adjustbox}{width=0.9\textwidth}
    \begin{tabular}{lcccccc}
      \hline

      System                             & Programming Language & Fully Neural & \begin{tabular}[c]{@{}c@{}}{State-of-the-art} \\{Performance}\end{tabular} & \begin{tabular}[c]{@{}c@{}}{Rich Chinese} \\{Fundamental Tasks}\end{tabular} & \begin{tabular}[c]{@{}c@{}}{Multi-task} \\{Learning}\end{tabular} \\ \hline
      LTP \cite{che-etal-2010-ltp}       & C++                  &              &                               & $\surd$                       &                               \\
      UDPipe  \cite{udpipe:2017}         & C++                  &              & $\surd$                       &                               &                               \\
      FLAIR \cite{akbik-etal-2019-flair} & Python               & $\surd$      & $\surd$                       &                               &                               \\
      Stanza \cite{qi2020stanza}         & Python               & $\surd$      & $\surd$                       &                               &                               \\
      \hline
      \texttt{N-LTP}                     & Python               & $\surd$      & $\surd$                       & $\surd$                       & $\surd$                       \\ \hline
    \end{tabular}
  \end{adjustbox}
  \caption{Feature comparisons of \texttt{N-LTP} against other popular natural language processing toolkits.}
  \label{comparisons}
\end{table*}
\section{Introduction}
There is a wide of range of existing natural language processing (NLP) toolkits such as CoreNLP~\cite{manning-etal-2014-stanford}, UDPipe \cite{udpipe:2017}, FLAIR \cite{akbik-etal-2019-flair}, spaCy,\footnote{\url{https://spacy.io}} and Stanza \cite{qi2020stanza} in English, which makes it easier for users to build tools with sophisticated linguistic processing.
Recently, the need for Chinese NLP has a dramatic increase in many downstream applications.
A Chinese NLP platform usually includes lexical analysis (Chinese word segmentation (CWS), part-of-speech (POS) tagging, and named entity recognition (NER)), syntactic parsing (dependency parsing (DEP)), and semantic parsing (semantic dependency parsing (SDP) and semantic role labeling (SRL)).
Unfortunately, there are relatively fewer high-performance and high-efficiency toolkits for Chinese NLP tasks.
To fill this gap, it's important to build a Chinese NLP toolkit to support rich Chinese fundamental NLP tasks, and make researchers process NLP tasks in Chinese quickly.

\begin{figure}[t]
  \centering
  \includegraphics[width=\linewidth]{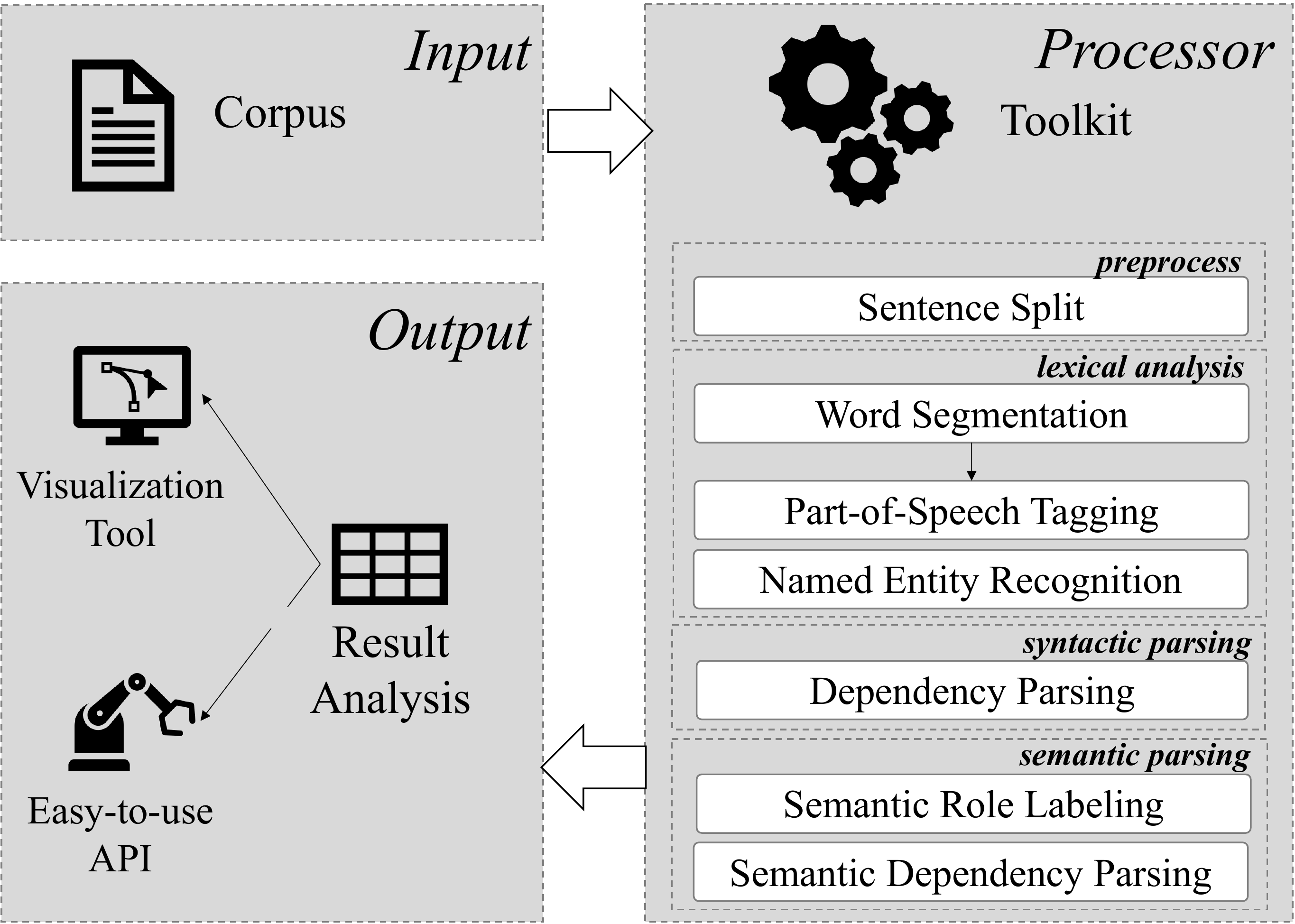}
  \caption{Workflow of the \texttt{N-LTP}.
    \texttt{N-LTP} takes the Chinese corpus as input and output the analysis results including lexical analysis, syntactic parsing, and semantic parsing.
    In addition, we provide the visualization tool and easy-to-use API to help users easily use \texttt{N-LTP}.
  }
  \label{architecture}
\end{figure}
Recently, \newcite{qi2020stanza} introduce the Python NLP toolkit \texttt{Stanza} for multi-lingual languages, including Chinese language.
Though \texttt{Stanza} can be directly applied for processing the Chinese texts, it suffers from several limitations.
First, it only supports part of Chinese NLP tasks. For example, it fails to handle semantic parsing analysis, resulting in incomplete analysis in Chinese NLP.
Second, it trained each task separately, ignoring the shared knowledge across the related tasks, which has been proven effective for Chinese NLP tasks \cite{qian-etal-2015-transition,hsieh-etal-2017-monpa,chang2018public}.
Third, independent modeling method will occupy more memory with the increase of the number of tasks, which makes it hard to deploy for mobile devices in real-word scenario.

To address the aforementioned issues, we introduce \texttt{N-LTP}, a PyTorch-based neural natural language processing toolkit for Chinese NLP, which was built on the SOTA pre-trained model.
As shown in Figure~\ref{architecture}, given Chinese corpus as input, \texttt{N-LTP} produces comprehensive analysis results, including lexical analysis, syntactic parsing, and semantic parsing. In addition, \texttt{N-LTP} provides easy-to-use APIs and visualization tool, which  is user-friendly.

As shown in Table~\ref{comparisons}, compared to the existing widely-used NLP toolkits, \texttt{N-LTP} has the following advantages:
\begin{itemize}

  \item \textbf{Comprehensive Tasks.} \texttt{N-LTP} supports rich Chinese fundamental NLP tasks including lexical analysis (word segmentation, part-of-speech tagging, named entity recognition), syntactic parsing, and semantic parsing (semantic dependency parsing, semantic role labeling).
        To the best of our knowledge, this is the first neural Chinese toolkit that support six Chinese fundamental NLP tasks.

  \item \textbf{Multi-Task Learning.}
        The existing NLP toolkits for the Chinese language all adopt independent models for each task, which ignore the shared knowledge across tasks.

        To alleviate this issue, we propose to use the multi-task framework \cite{collobert2011natural} to take advantage of the shared knowledge across all tasks.
        Meanwhile, multi-task learning with a shared encoder for all six tasks can greatly reduce the occupied memory and improve the speed, which makes \texttt{N-LTP}  more efficient, reducing the need for hardware.

        In addition, to enable the multi-task learning to enhance each subtask performance, we follow \newcite{DBLP:journals/corr/abs-1907-04829} to adopt the distillation method single-task models teach a multi-task
        model, helping the multi-task model surpass its all single-task teachers.  

  \item \textbf{Extensibility.} \texttt{N-LTP}  works with users’ custom modules. Users can easily add a new pre-trained model with a configuration file, in which users can change the pretrained model to any BERT-like model supported by HuggingFace Transformers \cite{Wolf2019HuggingFacesTS} easily by changing the config.
        We have made all task training configuration files open-sourced.
  \item \textbf{Easy-to-use API and Visualization Tool.}
        \texttt{N-LTP} provides a collection of fundamental APIs, which is convenient for users to use the toolkit without the need for any knowledge.
        We also provide a visualization tool, which enables users to view the processing results directly.
        In addition, \texttt{N-LTP} has bindings for many programming languages (C++, Python, Java, Rust, etc.).
  \item \textbf{State-of-the-art Performance.} We evaluate
        \texttt{N-LTP} on a total of six Chinese NLP tasks, and find that it achieves state-of-the-art or competitive performance at each task.

\end{itemize}

\texttt{N-LTP} is fully open-sourced and can support six Chinese fundamental NLP tasks. We hope \texttt{N-LTP} can facilitate Chinese NLP research.

\section{Design and Architecture}
\begin{figure}[t]
  \centering
  \includegraphics[width=\linewidth]{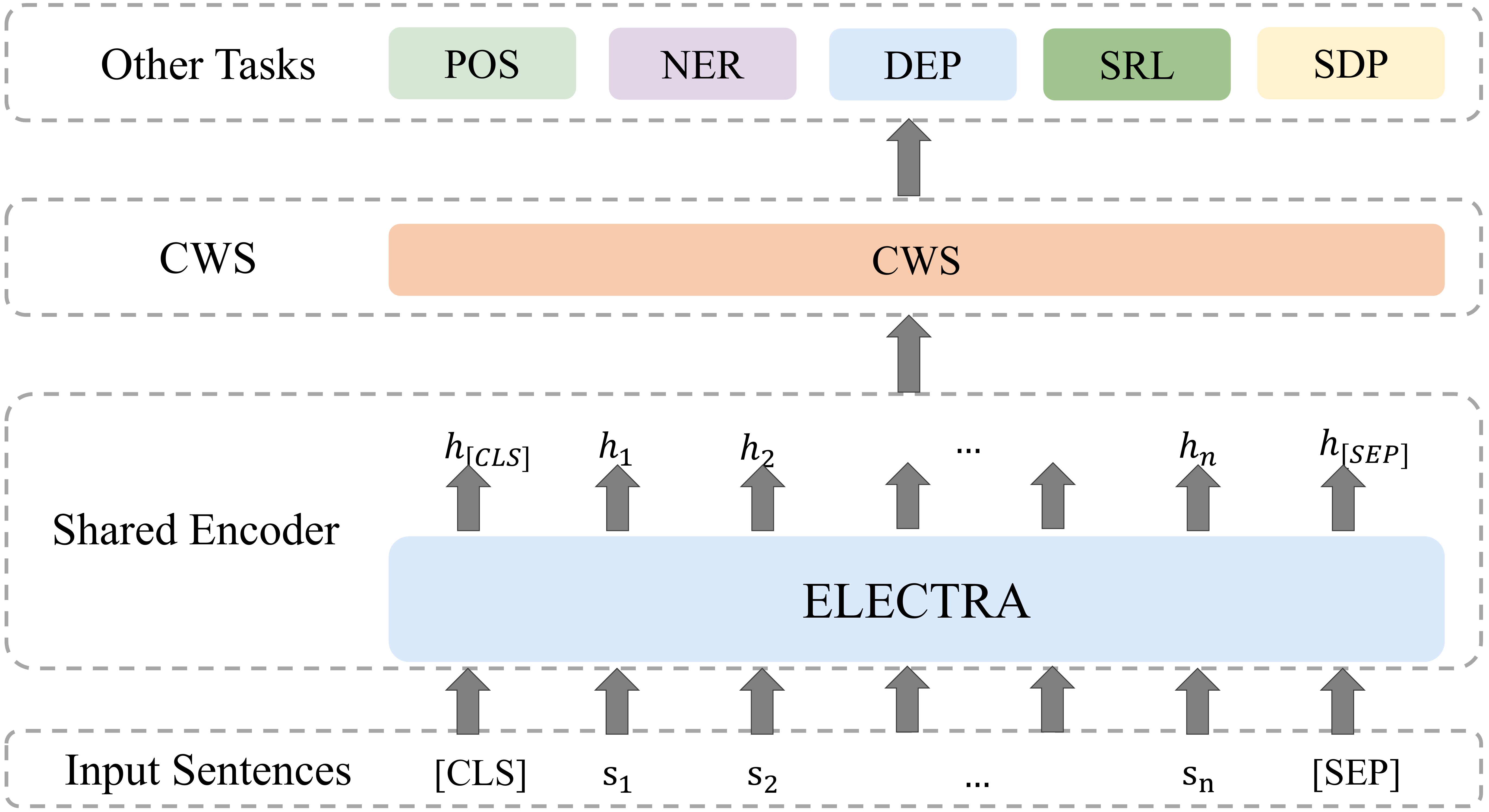}
  \caption{The architecture of the proposed model.}
  \label{model3}
\end{figure}

Figure~\ref{model3} shows an overview of the main architecture of \texttt{N-LTP}.
It mainly consists of the components including a shared encoder and different decoders for each task.
Our framework shares one encoder for leveraging the shared knowledge across all tasks.
Different task decoders are used for each task separately.
All tasks are optimized simultaneously via a joint learning scheme.
In addition, the knowledge distillation technique is introduced to encourage the multi-task model to surpass its single-task teacher model.

\subsection{Shared Encoder}
Multi-task framework uses a shared encoder to extract the shared knowledge across related tasks, which has obtained remarkable success on various NLP tasks \cite{qin-etal-2019-stack,wang-etal-2020-multi,zhou-etal-2021-end}.
Inspired by this, we adopt the SOTA pre-trained model (ELECTRA) \cite{clark2020electra} as the shared encoder to capture shared knowledge across six Chinese tasks.

Given an input utterance $s$ = ($s_{1}, s_{2}, \dots, s_{n}$), we first construct the input sequence by adding specific tokens $s$ = ($\texttt{[CLS]}, s_{1}, s_{2}, \dots, s_{n}, [\texttt{SEP}]$), where $\texttt{[CLS]}$ is the special symbol for representing the whole sequence, and $\texttt{[SEP]}$ is the special symbol to separate non-consecutive token sequences \cite{devlin2018bert}.
ELECTRA takes the constructed input and output the corresponding hidden representations of sequence $\boldsymbol{H}$ = ($\boldsymbol{h}_{\texttt{[CLS]}}$, $\boldsymbol{h}_{1}, \boldsymbol{h}_{2}, \dots, \boldsymbol{h}_{n}$, $\boldsymbol{h}_{\texttt{[SEP]}}$).

\subsection{Chinese Word Segmentation}
Chinese word segmentation (CWS) is a preliminary and important task for Chinese natural language processing (NLP).
In \texttt{N-LTP}, following \newcite{xue-2003-chinese}, CWS is regarded as a character based sequence labeling problem.

Specifically, given the hidden representations $\boldsymbol{H}$ = ($\boldsymbol{h}_{\texttt{[CLS]}}$, $\boldsymbol{h}_{1}, \boldsymbol{h}_{2}, \dots, \boldsymbol{h}_{n}$, $\boldsymbol{h}_{\texttt{[SEP]}}$), we adopt a linear decoder to classify each character:
\begin{eqnarray}
  \boldsymbol{y_{i}} = \mathrm {Softmax} (\boldsymbol{W}_{\text{CWS}} \boldsymbol{h_{i}} + \boldsymbol{b}_{\text{CWS}}),
\end{eqnarray}
where $	 \boldsymbol{y_{i}} $ denotes the label probability distribution of each character; $\boldsymbol{W}_{\text{CWS}} $ and $\boldsymbol{b}_{\text{CWS}} $ are trainable parameters.

\subsection{POS Tagging}
Part-of-speech (POS) tagging is another fundamental
NLP task, which can facilitate the downstream tasks such
as syntactic parsing.
Following the dominant model in the literature \cite{Ratnaparkhi1996AME,DBLP:journals/corr/HuangXY15}, POS tagging can be treated as a sequence labeling task.

Similar to CWS, we take the sequence of hidden representations $\boldsymbol{H}$ as input and output the corresponding POS sequence labels, which is formulated as:
\begin{eqnarray}
  \boldsymbol{y_{i}} = \mathrm {Softmax} ( \boldsymbol{W}_{\text{POS}}  \boldsymbol{h_{i}} +  \boldsymbol{b}_{\text{POS})},
\end{eqnarray}
where $	 \boldsymbol{y_{i}} $ denotes the POS label probability distribution of the $i$-th character; $	 \boldsymbol{h_{i}} $ is the ﬁrst sub-token representation of word $s_{i}$.

\subsection{Named Entity Recognition}
The named entity recognition (NER) is the task of
finding the start and end of an entity (\texttt{people}, \texttt{locations}, \texttt{organizations}, etc.) in a sentence
and assigning a class for this entity.

Traditional, NER is regarded as a sequence labeling task.
After obtaining the hidden representations $\boldsymbol{H}$, we follow \newcite{yan2019tener} to adopt the \texttt{Adapted-Transformer} to consider direction- and distance-aware  characteristic, which can be formulated as:
\begin{eqnarray}
  \hat{\boldsymbol{h_{i}}} = \mathrm{AdaptedTransformer} (\boldsymbol{h}_{i}),
\end{eqnarray}
where $\hat{\boldsymbol{H}}$ = ($\hat{\boldsymbol{h}}_{\texttt{[CLS]}}$, $\hat{\boldsymbol{h}}_{1}, \hat{\boldsymbol{h}}_{2}, \dots, \hat{\boldsymbol{h}}_{n}$, $\hat{\boldsymbol{h}}_{\texttt{[SEP]}}$) are the updated representations.

Finally, similar to CWS and POS, we use a linear decoder to classify label for each word:
\begin{eqnarray}
  \boldsymbol{y_{i}} = \mathrm {Softmax} (\boldsymbol{W}_{\text{NER}} \hat{\boldsymbol{h}_{i}} + \boldsymbol{b}_{\text{NER}}),
\end{eqnarray}
where $\boldsymbol{y_{i}}$ denotes the NER label probability distribution of each character.

\subsection{Dependency Parsing}\label{sec:dep}
Dependency parsing is the task
to analyze the semantic structure of a sentence.
In \texttt{N-LTP}, we implement a deep biaffine neural dependency parser \cite{deepbiaffine2017} and einser algorithm \cite{eisner-1996-three} to obtain the parsing result, which is formulated as:

\begin{eqnarray}\label{eq:biaffine}
  \begin{aligned}
    \boldsymbol{r}_i^{(\text{head})} & = & \text{MLP}^{(\text{head})}(\boldsymbol{h}_i) \\
    \boldsymbol{r}_j^{(\text{dep})}  & = & \text{MLP}^{(\text{dep})}(\boldsymbol{h}_j)
  \end{aligned}
\end{eqnarray}

After obtaining $ \boldsymbol{r}_i^{(\text{head})}$ and $  \boldsymbol{r}_j^{(\text{dep})}$,
we compute the score for each dependency $i^{\overset{\text{}}{\curvearrowleft}}j$ by:
\begin{equation}
  \bm{y}_{i^{\overset{\text{}}{\curvearrowleft}}j} = \text{BiAffine}(\bm{r}_i^{\text{dep}}, \bm{r}_j^{\text{head}}).
\end{equation}

The above process is also used for scoring a labeled dependency $i^{\overset{\text{l}}{\curvearrowleft}}j$,
by extending the 1-dim vector $\bm{s}$ into $L$ dims, where $L$ is the total number of dependency labels.
\begin{figure}[t]
	\centering
	\includegraphics[width=\linewidth]{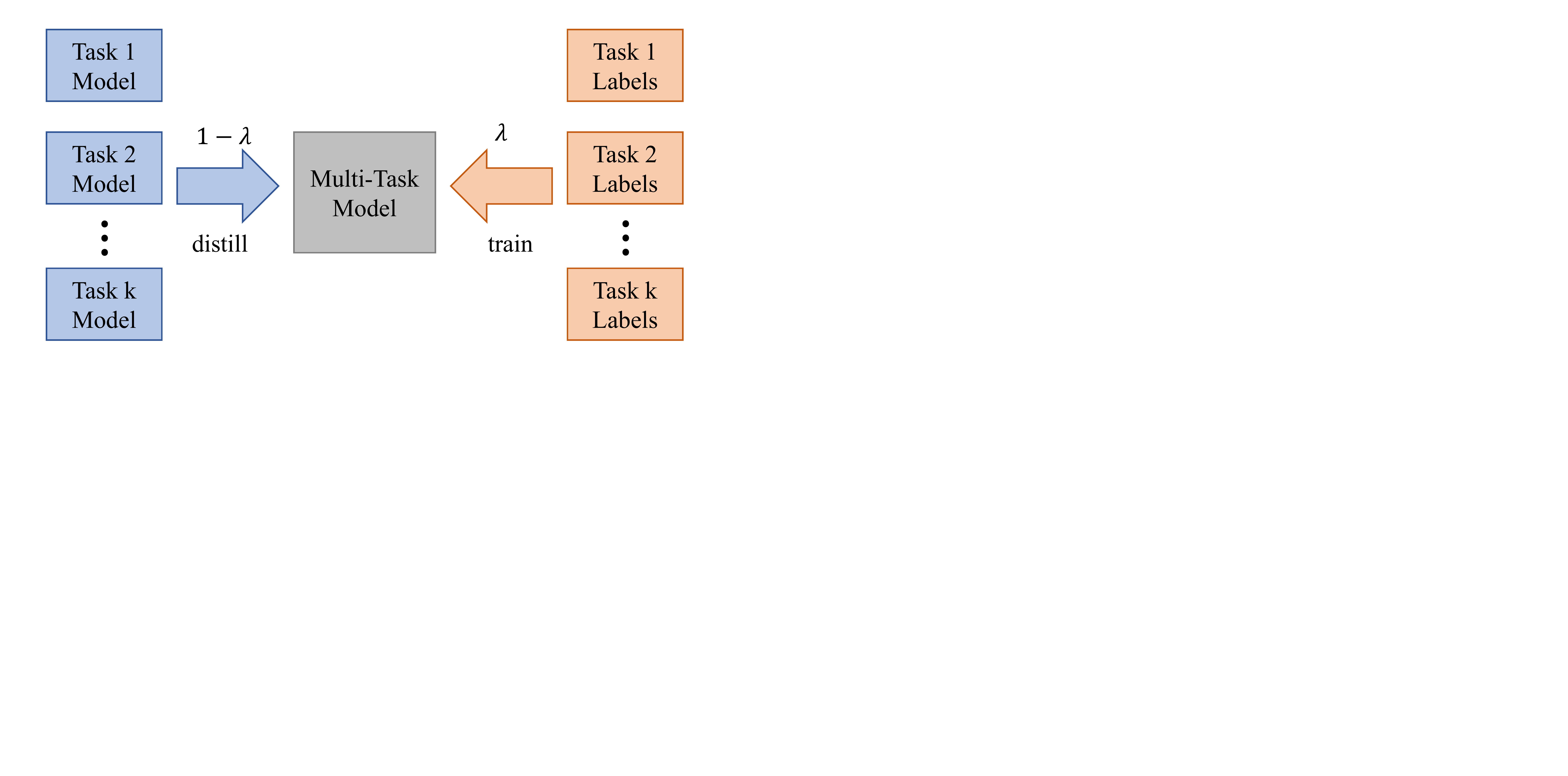}
	\caption{We follow \newcite{DBLP:journals/corr/abs-1907-04829} to adopt the distillation method. This is an overview of the distillation method. $\lambda$ is increased linearly from 0 to 1 over the curriculum of training.}
	\label{distill}
\end{figure}
\subsection{Semantic Dependency Parsing} Similar to dependency parsing,
semantic dependency parsing \cite[SDP]{che2016semeval} is a task to capture the semantic structure of a sentence.
Specifically, given an input sentence, SDP aims at determining all the word pairs
related to each other semantically and assigning specific predefined semantic relations.
Following \newcite{deepbiaffine2017}, we adopt a biaffine module to perform the task, using
\begin{eqnarray}\label{eq:sdp}
  \bm{p}_{i^{\overset{\text{}}{\curvearrowleft}}j} = \mathrm{sigmoid}(\bm{y}_{i^{\overset{\text{}}{\curvearrowleft}}j}).
\end{eqnarray}

If $\bm{p}_{i^{\overset{\text{}}{\curvearrowleft}}j} > 0.5$, word$_i$ to word$_j$ exists an edge.

\subsection{Semantic Role Labeling} Semantic Role Labeling (SRL) is the task of determining the latent predicate-argument structure of a sentence, which can provide representations to answer basic questions about sentence meaning, including who did what to whom, etc. We adopt an end-to-end SRL model by combining a deep biaffine neural network and a conditional random field (CRF)-based decoder \cite{cai-etal-2018-full}.

\begin{table*}[t]
  \centering
  \begin{adjustbox}{width=1.0\textwidth}
    \begin{tabular}{ccccccc}
      \toprule
                                        & \textbf{Chinese Word} & \textbf{Part-of-Speech} & \textbf{Named Entity} & \textbf{Dependency} & \textbf{Semantic Dependency} & \textbf{Semantic Role} \\
      \textbf{Model}                    & \textbf{Segmentation} & \textbf{Tagging}        & \textbf{Recognition}  & \textbf{Parsing}    & \textbf{Parsing}             & \textbf{Labeling}      \\

                                        & $F$                   & $F_{LAS}$               & $F$                   & $F$                 & $F$                          & $F$                    \\
      \midrule
      Stanza  \cite{qi2020stanza}       & 92.40                 & 98.10                   & 89.50                 & 84.98               & -                            & -                      \\

      \midrule
      \texttt{N-LTP} trained separately & 98.55                 & 98.35                   & 95.41                 & 90.12               & 74.47                        & 79.23                  \\
      \texttt{N-LTP} trained jointly with distillation    & \textbf{99.18}        & \textbf{98.69}          & \textbf{95.97}        & \textbf{90.19}      & \textbf{76.62}               & \textbf{79.49}         \\
      \bottomrule
    \end{tabular}
  \end{adjustbox}
  \caption{Main Results. ``-" represents the absence of tasks in the \texttt{Stanza} toolkit and we cannot report the results.}
  \label{main_result}
\end{table*}
The biaffine module is similar to Section~\ref{sec:dep} and the CRF layer can be formulated as:

\begin{eqnarray}\label{eq:crf}
  P(\hat{{y}}| s)=&\frac{\sum_{j=1}\exp f(y_{i,j-1},y_{i,j},s)}{\sum_{y_{i}'}\sum_{j=1}\exp f(y_{i,j-1}',y_{i,j}',s)}
\end{eqnarray}
where $\hat{y}$ represents an arbitrary label sequence when predicate is $s_i$, and $f(y_{i,j-1},y_j,s)$ computes the transition score from $y_{i,j-1}$ to $y_{i,j}$.

\begin{figure}[t]
  \centering
  \includegraphics[width=\linewidth]{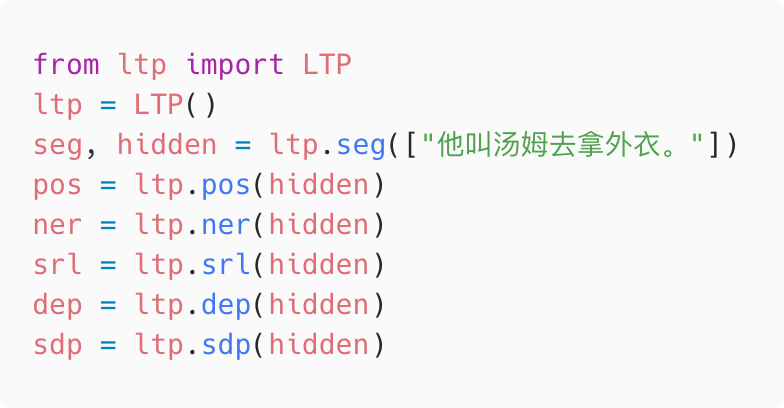}
  \caption{A minimal code snippet.}
  \label{code}
\end{figure}
\subsection{Knowledge Distillation}

When there exist a large number of tasks, it's difficult to ensure that each task task benefits from multi-task learning~\cite{DBLP:journals/corr/abs-1907-04829}.

Therefore,  we follow BAM \cite{DBLP:journals/corr/abs-1907-04829} to use the knowledge distillation to alleviate this issue, which is shown Figure~\ref{distill}.
First, we train each task as the teacher model.
Then, \texttt{N-LTP} learns from each trained single-task teacher model while learning from the gold-standard labels simultaneously.

Following BAM \cite{DBLP:journals/corr/abs-1907-04829},
we adopt teacher annering distillation algorithm.
More specifically, instead of simply shuffling the datasets for our multi-task models, we follow the task sampling procedure from \newcite{DBLP:journals/corr/abs-1812-10860}, where the probability of training on an example for a particular task $\tau$ is proportional to $|D\tau|^{0.75}$.  This ensures that tasks with  large datasets don’t overly dominate the training.

\section{Usage}
\texttt{N-LTP} is a PyTorch-based Chinese NLP toolkit based on the above model.
All the conﬁgurations can be initialized from JSON ﬁles, and thus it is easy for users to use \texttt{N-LTP} where users just need one line of code to load the model or process the input sentences. Specifically, \texttt{N-LTP} can be installed easily by the command:

\begin{lstlisting}[language=Bash]
  $ pip install ltp
\end{lstlisting}

In addition, \texttt{N-LTP} has bindings available for many programming languages, including C++, Python, Java and RUST directly.

\subsection{Easy-to-use API}
We provide rich easy-to-use APIs, which enables users to easily use without the need for any knowledge.
The following code snippet in Figure~\ref{code} shows a minimal usage of \texttt{N-LTP} for downloading models, annotating a sentence with customized models, and predicting all annotations.

\subsection{Visualization Tool}
In addition, a visualization tool is proposed   for users to view the processing results directly.
Specifically, we build an interactive web demo that runs the pipeline interactively, which is publicly available at \url{http://ltp.ai/demo.html}.
\begin{figure}[t]
  \centering
  \includegraphics[width=0.95\columnwidth]{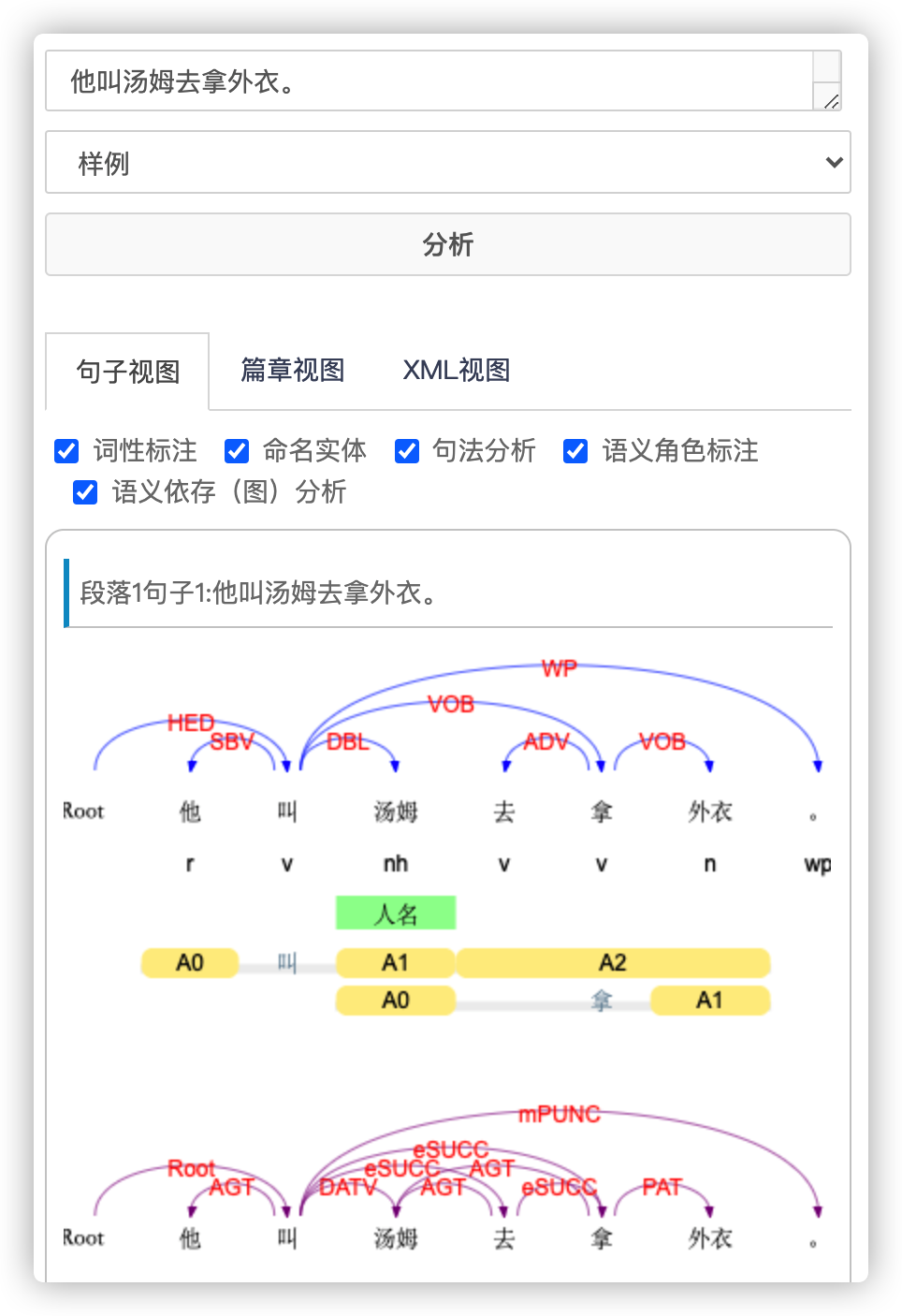}
  \caption{LTP annotates a Chinese sentence ``\begin{CJK*}{UTF8}{gbsn}他叫汤姆去拿外衣。\end{CJK*}/ He told Tom to get his coat.''. The output is visualized by our visualization  demo.}
  \label{demo}
\end{figure}
The visualization tool is shown in Figure~\ref{demo}.

\section{Experiments}

\begin{table*}[htbp]
  \centering
  \resizebox{1.0\textwidth}{!}{%
    \begin{tabular}{crrcccc}
      \toprule
      \multirow{2}{*}{Task} & \multirow{2}{*}{Model}                                     & \multirow{2}{*}{Dataset} & \multirow{2}{*}{Metric} & State-of-the-art & \texttt{N-LTP}     & \texttt{N-LTP}  \\
                            &                                                            &                          &                         & Performance      & trained separately & trained jointly \\ \midrule
      CWS                   & BERT~~\cite{DBLP:journals/corr/abs-1903-04190}             & 10 Corpus\footnote{10 corpus for training CWS task includes PKU, MSR, AS, CITYU, XU, CTB, UDC, CNC, WTB and ZX.}  & $\mathrm{F}1$           & 97.10            & 97.42              & \textbf{97.50}  \\
      POS                   & Glyce+BERT~~\cite{DBLP:journals/corr/abs-1901-10125}       & CTB9                     & $\mathrm{F}1$           & 93.15            & 94.57              & \textbf{95.17}  \\
      NER                   & ZEN~~\cite{DBLP:journals/corr/abs-1911-00720}              & MSRA                     & $\mathrm{F}1$           & 95.25            & 94.95              & \textbf{95.78}  \\
      NER                   & DGLSTM-CRF~~\cite{DBLP:journals/corr/abs-1909-10148}       & OntoNotes                & $\mathrm{F}1$           & 79.92            & 84.08              & \textbf{84.38}  \\
      SRL                   & BiLSTM-Span~~\cite{D18-1191}                               & CONLL12                  & $\mathrm{F}1$           & 75.75            & 78.20              & \textbf{81.65}  \\
      DEP                   & Joint-Multi-BERT~~\cite{DBLP:journals/corr/abs-1904-04697} & CTB9                     & $\mathrm{F}1_{LAS}$     & 81.71            & 81.69              & \textbf{84.03}  \\
      SDP                   & SuPar\footnote{\url{http://ir.hit.edu.cn/sdp2020ccl}}                                    & CCL2020\footnote{\url{http://ir.hit.edu.cn/sdp2020ccl}}  & $\mathrm{F}1_{LAS}$     & \textbf{80.38}   & 76.27              & 75.76           \\  \bottomrule
    \end{tabular}%
  }
  \caption{The results of \texttt{N-LTP} comparation to other state-of-the-art performance..}
  \label{tab:sota-compare}
\end{table*}

\subsection{Experimental Setting}
To evaluate the efficiency of our multi-task model, we conduct experiments on six Chinese tasks.

The \texttt{N-LTP} model is based on the Chinese ELECTRA base \cite{cui-etal-2020-revisiting}.
The learning ratio (lr) for teacher models, student model
and CRF layer is $\{1e-4\}$, $\{1e-4\}$, $\{1e-3\}$, respectively.
The gradient clip value adopted in our experiment is 1.0 and the warmup proportion is 0.02.
We use BertAdam \cite{devlin2018bert} to optimize the parameters and adopted the suggested hyper-parameters for optimization.


\subsection{Results}

We compare \texttt{N-LTP} with the state-of-the-art toolkit \texttt{Stanza}.
For a fair comparison, we conduct experiments on the same datasets that \texttt{Stanza} adopted.

The results are shown in Table~\ref{main_result}, we have the following observations:
\begin{itemize}
  \item \texttt{N-LTP} outperforms \textit{Stanza} on four common tasks including CWS, POS, NER, and DEP by a large margin, which shows the superiority of our proposed toolkit.
  \item  The multi-task learning outperforms the model with independently trained.
        This is because that the multi-task framework can consider the shared knowledge which can promote each task compared with the independently training paradigm.
\end{itemize}

\subsection{Analysis}

\subsubsection{Speedup and Memory Reduction}
In this section,
we perform the speed and memory test on the Tesla V100-SXM2-16GB and
all models were speed-tested on the 10,000 sentences of the People's Daily corpus with a batch size of 8.
In all experiments, \texttt{N-LTP} performs six tasks (CWS, POS, NER, DEP, SDP, SRL) while \texttt{Stanza} only conduct four tasks (CWS, POS, NER, DEP).

\begin{figure}[t]
  \centering
  \includegraphics[width=0.95\linewidth]{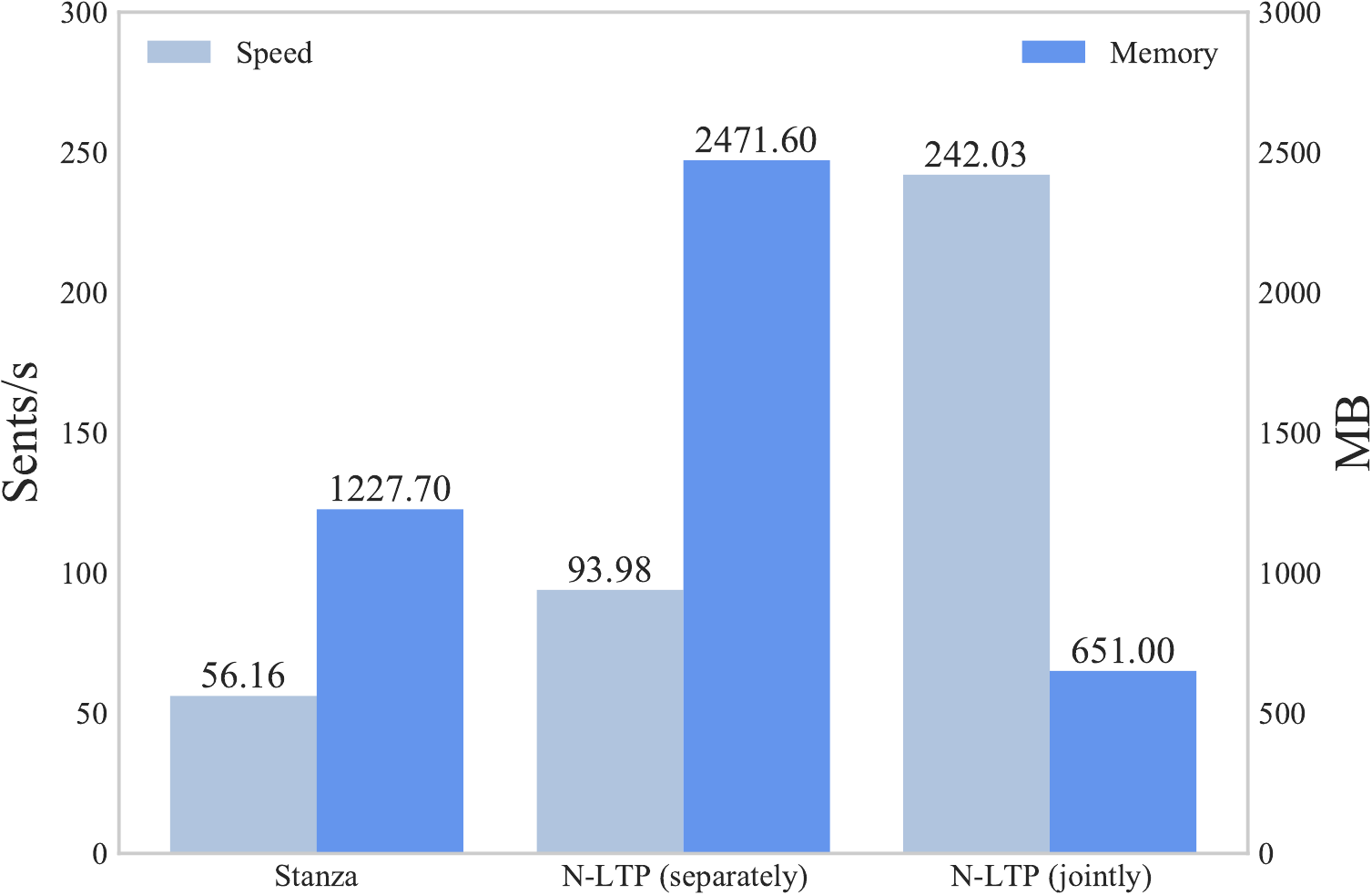}
  \caption{Speed and Memory test for \texttt{N-LTP}.}\label{compare-speed}
\end{figure}

\footnotetext[2]{10 corpus for training CWS task includes PKU, MSR, AS, CITYU, XU, CTB, UDC, CNC, WTB and ZX.\label{corpus}}
\footnotetext[3]{\url{http://ir.hit.edu.cn/sdp2020ccl} \label{ccl2020}}
\paragraph{Speedup}
We compare the speed between \texttt{Stanza}, \texttt{N-LTP-separately} and \texttt{N-LTP-jointly} and the results are shown in Figure~\ref{compare-speed}.
From the results of speed test, we have two interesting observations: 
  (1)\texttt{N-LTP trained separately} achieves the {x1.7} speedup compared with \texttt{Stanza}. We attribute that \texttt{N-LTP} adopts the transformer as an encoder that can be calculated in parallel while \texttt{Stanza} uses LSTM which can only process sentences word by word; (2) \texttt{N-LTP trained jointly with distillation} obtains the {x4.3} speedup compared with separate modeling paradigm. This is because that our model utilizes the multi-task to perform all tasks while the independent models can be only processed all tasks in a pipeline mode.

\paragraph{Memory Reduction}
For memory test, we have the following observation:
(1) \texttt{N-LTP trained separately} occupy more memory than \texttt{Stanza}.
This is because \texttt{N-LTP} performs six tasks while \texttt{Stanza} only conduct four tasks.
(2) Though performing six tasks, \texttt{N-LTP trained jointly} only requires half the memory compared to \texttt{Stanza}.
We attribute it to the fact that the multi-task framework with a shared encoder can greatly reduce the running memory.

\subsubsection{Comparation with Other SOTA Single Models}
To further verify the effectiveness of \texttt{N-LTP}, we compare our framework with the existing state-of-the-art single models on six Chinese fundamental tasks.
In this comparison, we conduct experiments on the same wildly-used dataset in each task for a fair comparison.
In addition, we use \texttt{BERT} rather than \texttt{ELECTRA} as the shared encoder, because the prior work adopts \texttt{BERT}.

Table \ref{tab:sota-compare} shows the results, we observe that our framework obtains best performance on five out of six tasks including CWS, POS, NER, SRL, and DEP, which demonstrates the effectiveness of our framework.
On the SDP task, \texttt{N-LTP} underperforms the best baseline. This is because many tricks are used in the prior model for SDP task and we just use the basic multi-task framework.

\section{Conclusion}
In this paper, we presented \texttt{N-LTP},
an open-source neural language technology platform supporting Chinese. To the best of our knowledge, this is the first Chinese toolkit that supports six fundamental Chinese NLP tasks.
Experimental results show \texttt{N-LTP} obtains state-of-the-art or competitive performance and has high speed..
We hope \texttt{N-LTP} can facilitate Chinese NLP research.
\section*{Acknowledgements}
We thank the anonymous reviewers for their detailed and constructive comments. The ﬁrst three authors contributed equally. Wanxiang Che is the corresponding author. This work was supported by the National Key R\&D Program of China via grant 2020AAA0106501 and
the National Natural Science Foundation of China
(NSFC) via grant 61976072 and 61772153. Libo is also supported by the Zhejiang Lab’s
International Talent Fund for Young Professionals.

\bibliography{anthology,acl2021}
\bibliographystyle{acl_natbib}

\end{document}